\documentclass[10pt,twocolumn,letterpaper]{article}

\usepackage{iccv}
\usepackage{times}
\usepackage{epsfig}
\usepackage{graphicx}
\usepackage{amsmath}
\usepackage{amssymb}

\usepackage{amsfonts}
\usepackage{algorithmic}
\usepackage{textcomp}
\usepackage{subfigure}
\usepackage{array}
\usepackage[table]{xcolor}
\usepackage{microtype}
\usepackage{nameref}
\usepackage{chngpage}
\usepackage{multirow}
\usepackage{booktabs}  
\usepackage{threeparttable} 

\usepackage[breaklinks=true,bookmarks=false]{hyperref}

\iccvfinalcopy 

\def\iccvPaperID{****} 

\ificcvfinal\fi

\begin{document}

\title{Multi Voxel-Point Neurons Convolution (MVPConv) for Fast and Accurate 3D Deep Learning}

\author{Wei Zhou\\
Northwest University\\
{\tt\small mczhouwei12@gmail.com}
\and
Xin Cao\\
Northwest University\\
{\tt\small xin\_cao@163.com}
\and
Xiaodan Zhang\\
Northwest University\\
{\tt\small xiaodanzhang@nwu.edu.cn}
\and
Xingxing Hao\\
Northwest University\\
{\tt\small ystar1991@126.com}
\and
Dekui Wang\\
Northwest University\\
{\tt\small dekui\_wang@126.com}
\and
Ying He\\
Nanyang Technological University\\
{\tt\small yhe@ntu.edu.sg}
}

\maketitle
\ificcvfinal\thispagestyle{empty}\fi

\begin{abstract}
We present a new convolutional neural network, called Multi Voxel-Point Neurons Convolution (MVPConv), for fast and accurate 3D deep learning. 
The previous works adopt either individual point-based features or local-neighboring voxel-based features to process 3D model, which limits the performance of models due to the inefficient computation. Moreover, most of the existing 3D deep learning frameworks aim at solving one specific task, and only a few of them can handle a variety of tasks.
Integrating both the advantages of the voxel and point-based methods, the proposed MVPConv can effectively increase the neighboring collection between point-based features and also promote the independence among voxel-based features.
Simply replacing the corresponding convolution module with MVPConv, we show that MVPConv can fit in different backbones to solve a wide range of 3D tasks.  
Extensive experiments on benchmark datasets such as ShapeNet Part, S3DIS and KITTI for various tasks show that MVPConv improves the accuracy of the backbone (PointNet) by up to \textbf{36\%}, and achieves higher accuracy than the voxel-based model with up to \textbf{34}$\times$ speedup. In addition, MVPConv  also outperforms the state-of-the-art point-based models with up to \textbf{8}$\times$ speedup. Notably, our MVPConv achieves better accuracy than the newest point-voxel-based model PVCNN (a model more efficient than PointNet)
with lower latency.
\end{abstract}

\section{Introduction}

3D deep learning for point clouds has received much attention in both industry and academia thanks to its potential for a wide range of applications, such as autonomous driving and robots.
The main technical challenges are due to the the sparse and irregular nature of point clouds. 

The existing 3D deep learning methods can be roughly divided into voxel- and point-based methods according to the representations of point clouds.
The voxel-based methods convert the irregular and sparse point clouds into regular 3D grids so that the widely studied convolutional neural networks (CNN) can be applied directly~\cite{cciccek20163d,riegler2017octnet,zhou2018voxelnet}.
Since their performance heavily depends on the voxelization resolution, the voxel-based methods often suffer from large information loss when the resolution is low, as multiple adjacent points are quantized into the same grid, which are indistinguishable. Conversely, a high resolution volume would preserve the fine-detailed information, but requires significant amount of GPU memory and computation time due to the cubic complexity of volumes. 
In contrast, the point-based methods can handle high-resolution models, since they process the raw points in a local and separate manner~\cite{klokov2017escape,Li2019pointcnn,qi2017pointnet,qi2017pointnetplusplus,wang2019dynamic}. Taking advantage of the sparse representation of point clouds, the point-based methods consume much less GPU memory than the voxel-based methods.  However, due to lack of regularity, they suffer from expensive random memory access and dynamic kernel computation during the point and its nearest neighbor searching~\cite{liu2019pvcnn}.

Motivated by the merits and limitations of each type of methods, several researchers proposed mixed representations to overcome the challenges of high accuracy demand and limited computational resources available on GPUs recently ~\cite{liu2019pvcnn,cherenkova2020pvdeconv,shi2021pv,zhang2020deep,tang2020searching,shi2020pv}. However, most of these point-voxel combined methods are only for solving a specific task with a point-voxel based framework. For example,  PV-RCNN~\cite{shi2020pv} and PV-RCNN++~\cite{shi2021pv} focus on 3D object detection, and  Pvdeconv~\cite{cherenkova2020pvdeconv} targets 3D auto-encoding CAD construction, FusionNet~\cite{zhang2020deep} is for semantic segmentation. Tang \etal proposed a sparse point-voxel convolution for efficient 3D architecture searching~\cite{tang2020searching}. 
To our knowledge, there is no general point-voxel based method for solving different kind of tasks.

\begin{figure*}[t]
	\begin{center}
		\includegraphics[width=1\linewidth]{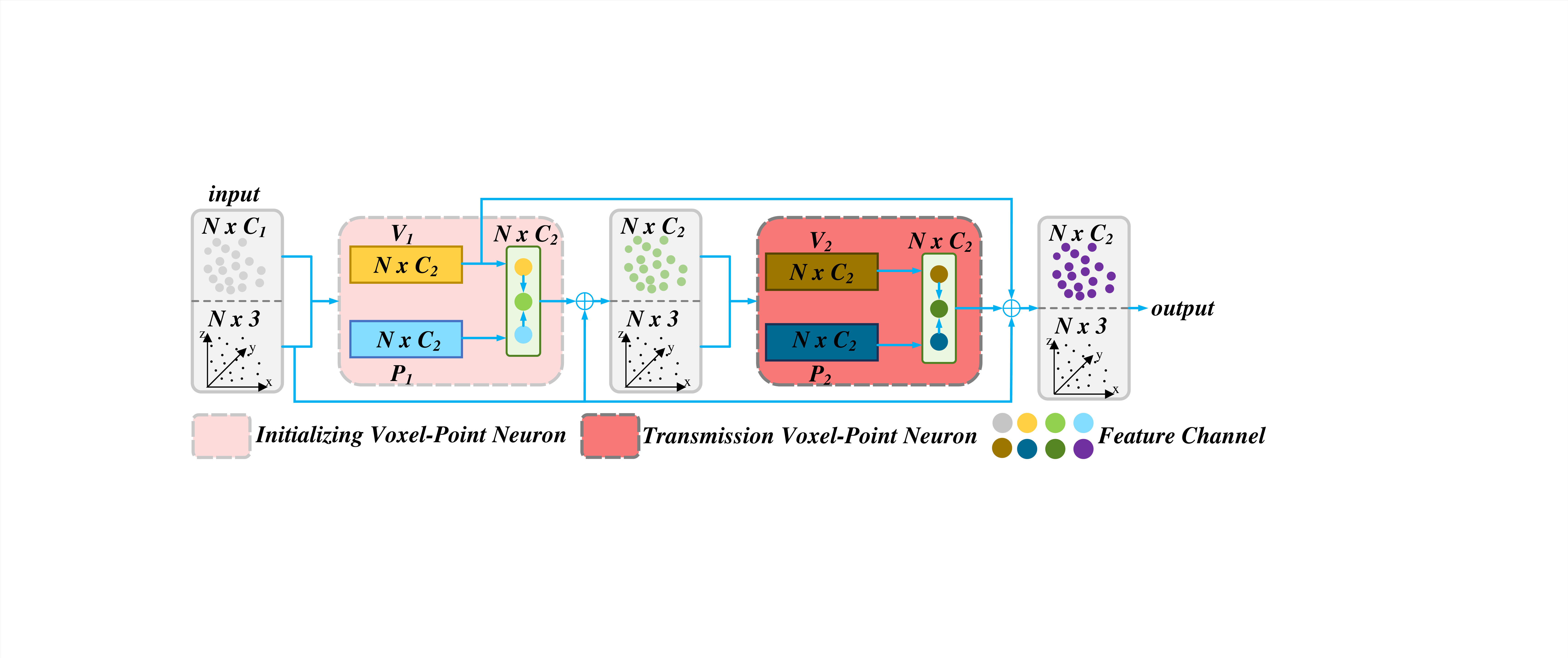}
	\end{center}
	\caption{The proposed MVPConv increases the neighboring collection between point-based features and the independence among voxel-based features via adopting both 3D CNN and MLPs on both point-based and voxel-based features.}
	\label{fig:mvpconv_simple}
\end{figure*}

To design effective deep neural networks for 3D analysis, one must take into consideration the performance, efficiency, as well as generality and flexibility for various tasks. In this paper, we propose Multi Voxel-Point Neurons Convolution (MVPConv) which takes the advantages of both the voxel- and point-based methods, and can work with different backbones for a wide range of 3D tasks (see  Figure~\ref{fig:mvpconv_simple}). 
In the previous 3D CNN models, the point-based features are individual and the voxel-based features are based on local neighborhood. In contrast, our MVPConv conducts 3D CNN and MLP on both points and voxels. As a result, it increases not only the neighboring collection for point-based features, but also the independence among voxel-based features. Extensive experiments show that MVPCNN outperforms the state-of-the-art 3D CNN models in terms of accuracy and efficiency.

\section{Related Work}
\label{sec:related_work}

\textbf{Voxel-based 3D learning.} Inspired by the success of CNN on 2D images~\cite{tran2015learning,ren2016faster,redmon2016you}, researchers transfer point cloud representation to volumetric representation and attempt to adopt convolution over it~\cite{maturana2015voxnet,cciccek20163d,qi2016volumetric,le2018pointgrid,du2020vipnet}. As is know to all, the larger the voxel resolution is, the more detailed feature information is contained, and the calculation of 3D volumetric convolution increases exponentially. To slow down this problem, researchers adopt octree to construct efficient convolutional architectures to increase the efficiency of computation with high voxel resolution~\cite{tatarchenko2017octree,riegler2017octnet,tatarchenko2017octree}.
State-of-the-art researches have demonstrated that volumetric representation can also be applied in 3D shape classification~\cite{wu20153d,wang2017cnn,le2018pointgrid}, 3D point cloud segmentation~\cite{graham20183d,meng2019vv,wang2019voxsegnet} and 3D object detection~\cite{zhou2018voxelnet}. Although the voxel-based methods have a great advantage in data structuring for 3D CNN, its computation efficiency is still greatly limited by the size of voxel resolution.

\textbf{Point-based 3D learning.} PointNet~\cite{qi2017pointnet}, the first deep neural network, takes advantage of spacial transform network (STN) and simple symmetric function (maxpooling) to process 3D point clouds. Many follow-up researches improves PointNet by aggregating  hierarchically to extract the local features~\cite{qi2017pointnetplusplus,klokov2017escape}. PointCNN~\cite{Li2019pointcnn}, SpiderCNN~\cite{xu2018spidercnn} and Geo-CNN~\cite{lan2019modeling} dynamically generate local geometric structure to capture points' neighboring features. RSNet~\cite{huang2018recurrent} adopts a lightweight local dependency module to efficiently model local structures of point clouds.
3D-GCN~\cite{lin2021learning}, DGCNN~\cite{wang2019dynamic}, Grid-GCN~\cite{xu2020grid}, SpecConv~\cite{wang2018local}, SPGraph~\cite{landrieu2018large}, GAC~\cite{wang2019graph} adopt graph convolutional networks to conduct 3D point cloud learning, while RS-CNN~\cite{liu2019relation}, PCNN~\cite{atzmon2018point}, SCN~\cite{xie2018attentional} and KCNet~\cite{shen2018mining} make use of geometric relations for point cloud analysis. Furthermore, SPLATNet~\cite{su2018splatnet} uses sparse bilateral convolutional layers to build the network, and SO-Net~\cite{li2018so} proposes permutation invariant architectures for learning with unordered point clouds. SSNet~\cite{thabet2020self} combines Morton-order curve and point-wise to conduct self-supervised learning. SPNet~\cite{liu2020self} uses a self-prediction for 3D instance and semantic segmentation of point clouds. As pointed out in~\cite{liu2019pvcnn}, random memory access and dynamic kernel computation are the performance bottleneck of point-based methods.

\textbf{Point-voxel-based 3D learning.} Point-voxel-based methods combine the voxel- and point-based approaches. Aiming at 3D object detection, PV-RCNN~\cite{shi2020pv} attracts the multi-scaled 3D voxel CNN features by PointNet++~\cite{qi2017pointnetplusplus}, then the voxel- and point-based features are aggregated to a small set of representative points. Based on PV-RCNN, PV-RCNN++~\cite{shi2021pv} has improved the computation efficiency of point-based parts. Like PV-RCNN and PV-RCNN++ applied in 3D object detection, other point-voxel-based 3D learning methods~\cite{cherenkova2020pvdeconv,zhang2020deep,tang2020searching} are specialized in 3D construction, 3D semantic segmentation or 3D structuring.
PVCNN~\cite{liu2019pvcnn} adopts interpolation to obtain the voxel-based CNN features, and then aggregates the point-based features and the interpolated voxel-based features as output. Comparing with the existing point-voxel-based methods~\cite{liu2019pvcnn,cherenkova2020pvdeconv,shi2021pv,zhang2020deep,tang2020searching,shi2020pv}, our method can deal with different kind of 3D tasks by applying MVPConv in various backbones. Moreover, we use shared MLPs to process the voxel-based features and adopt 3D CNN on the point-based features, thus improving the performances of different 3D deep learning tasks.

\section{Multi Voxel-Point Neurons Convolution}
\label{sec:mvpconv}

\begin{figure*}[t]
	\begin{center}
		\includegraphics[width=1\linewidth]{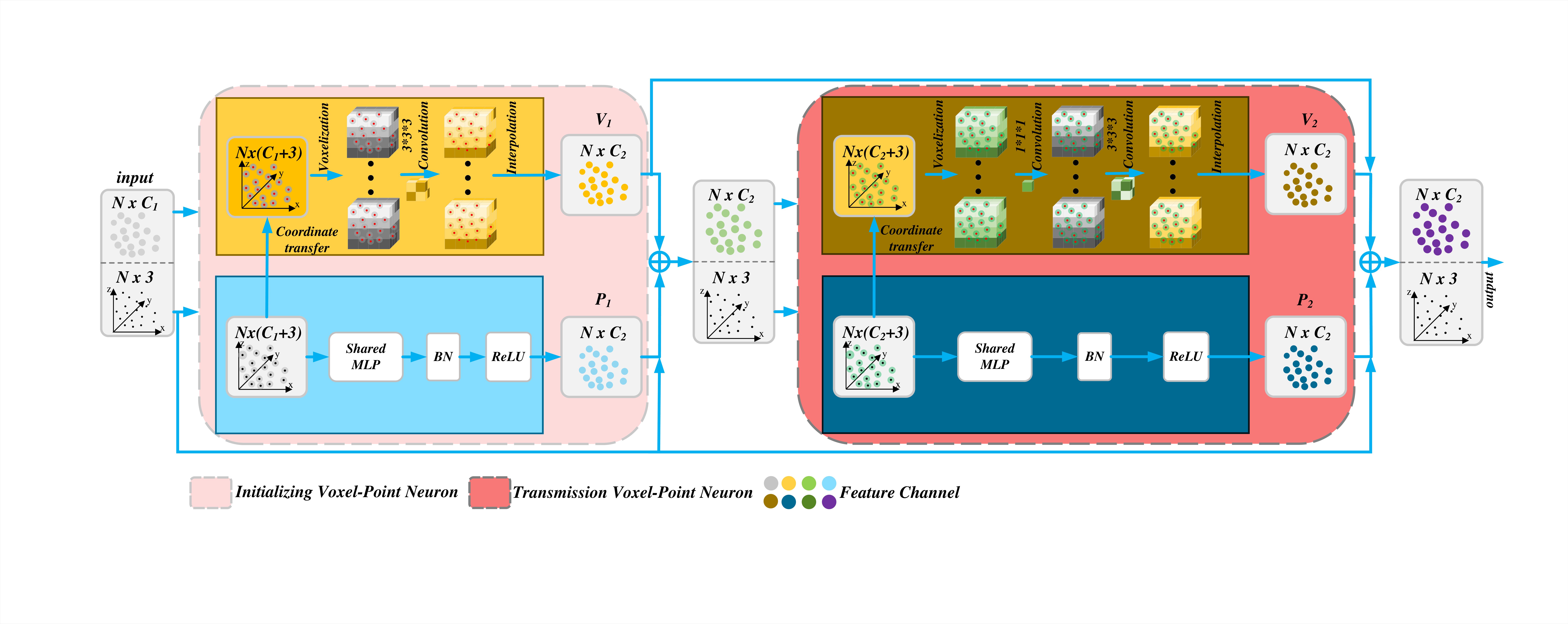}
	\end{center}
	\caption{MVPConv pipeline.}
	\label{fig:mvpconv}
\end{figure*}

State-of-the-art 3D deep learning methods are based on either voxel-based CNN methods or point-based network. Generally speaking, 3D voxel-based CNN method owns good data locality and regularity for low-resolution point neighbors (voxel grids), while the point-based approaches are independent for each point, so it could capture high-resolution information. 

In this paper, we propose a new way to realize convolution for processing point cloud that named Multi Voxel-Point Neurons Convolution (MVPConv) inspired by point-voxel-based methods and convolution operation. Instead of working on a specific 3D task like PV-RCNN~\cite{shi2020pv}, PV-RCNN++~\cite{shi2021pv}, Pvdeconv~\cite{cherenkova2020pvdeconv} and FusionNet~\cite{zhang2020deep}, our MVPConv could be applied in the backbone frameworks of different tasks (replacing the corresponding convolution in backbone with MVPConv). 
Furthermore, the previous point-voxel-based methods (e.g.~\cite{shi2020pv,shi2021pv,liu2019pvcnn}) conduct the 3D CNN and MLP operations for the point and voxel separately, and finally gather the point-based and voxel-based features together. Unlike them, our MVPConv performs both 3D CNN and MLP on both point and voxel.

As illustrated in Figure.~\ref{fig:mvpconv}, MVPConv consists of two Voxel-Point neurons, the left neuron is initializing neuron, the right one is the transmission neuron, each neuron can be divides into the point-based and voxel-based modules. 
The point-based and voxel-based modules in these two neurons got big similarities:
the point-based modules mainly use shared MLP to extract the independent features for each point, and it only consumes a small amount of memory for calculation even for high-resolution points; 
the voxel-based modules adopt 3D CNNs and interpolation to capture the neighboring points features, and it only cost a small GPU memory as the voxel resolution adopted in this paper is low.

\renewcommand{\arraystretch}{1.0}  
\renewcommand\tabcolsep{3.0pt}
\begin{table*}[ht]
	\centering  
	\fontsize{9}{11}\selectfont  
	\begin{threeparttable}  
		\caption{Evaluation results of part segmentation on ShapeNet Part dataset.}  
		\label{tab:shapenet}  
		\begin{tabular}{lccccccccccccccccccccccc}  
			\toprule  
			Method&Reference&Type&Input Data&mIoU&GPU Mem.&latency\cr  
			\midrule
			mIoU $<$ 86.0\cr
			Kd-Net~\cite{klokov2017escape}&ICCV 2017&point-based&$8 \times 4K$ points&82.3&-&-\cr
			PointNet~\cite{qi2017pointnet}&CVPR 2017&point-based&$8 \times 2K$ points&83.7&1.5GB&21.7ms\cr
			3D-UNet~\cite{cciccek20163d}&MICCAI 2016&voxel-based&$8 \times 96^3$ voxels&84.6&8.8GB&682.1ms\cr 	
			SO-Net~\cite{li2018so}&CVPR 2018&point-based&$8 \times 1K$ points&84.6&-&-\cr
			SCN~\cite{xie2018attentional}&CVPR 2018&point-based&$8 \times 1K$ points&84.6&-&-\cr
			SPLATNet~\cite{su2018splatnet}&CVPR 2018&point-based&-&84.6&-&-\cr	
			KCNet~\cite{shen2018mining}&CVPR 2018&point-based&$8 \times 2K$ points&84.7&-&-\cr
			RSNet~\cite{huang2018recurrent}&CVPR 2018&point-based&$8 \times 2K$ points&84.9&0.8GB&74.6ms\cr
			PointNet++~\cite{qi2017pointnetplusplus}&NeurIPS 2017&point-based&$8 \times 2K$ points&85.1&2.0GB&77.9ms\cr
			DGCNN~\cite{wang2019dynamic}&SIGGRAPH 2019&point-based&$8 \times 2K$ points&85.1&2.4GB&87.8ms\cr
			PCNN~\cite{atzmon2018point}&SIGGRAPH 2018&point-based&$8 \times 2K$ points&85.1&-&-\cr
			SpiderCNN~\cite{xu2018spidercnn}&ECCV 2018&point-based&$8 \times 2K$ points&85.3&6.5GB&170.7ms\cr
			GSNet~\cite{xu2020geometry}&AAAI 2020&point-based&-&85.3&-&-\cr
			3D-GCN~\cite{lin2021learning}&TPAMI 2021&point-based&$8 \times 2K$ points&85.3&-&-\cr
			$\textbf{MVPCNN}_{(0.25 \times Ch)}$&-&voxel-point-based&$8 \times 2K$ points&85.5&1.7GB&\textbf{19.8ms}\cr
			$\textbf{MVPCNN}_{(0.5 \times Ch)}$&-&voxel-point-based&$8 \times 2K$ points&\textbf{85.7}&2.1GB&31.0ms\cr
			\midrule
			mIoU $>$ 86.0\cr
			PointCNN~\cite{Li2019pointcnn}&NeurIPS 2018&point-based&$8 \times 2K$ points&86.1&2.5GB&135.8ms\cr
			SPNet~\cite{liu2020self}&ECCV 2020&point-based&$8 \times 2K$ points&86.2&-&-\cr
			CF-SIS~\cite{wen2020cf}&ACM MM 2020&point-based&$8 \times 2K$ points&86.2&-&-\cr
			PVCNN~\cite{liu2019pvcnn}&NeurIPS 2019&voxel-point-based&$8 \times 2K$ points&86.2&1.6GB&\textbf{50.7ms}\cr
			RS-CNN~\cite{liu2019relation}&CVPR2019&point-based&$8 \times 2K$ points&86.2&-&-\cr
			$\textbf{MVPCNN}_{(1 \times Ch)}$&-&voxel-point-based&$8 \times 2K$ points&\textbf{86.5}&2.8GB&81.9ms\cr
			\bottomrule  
		\end{tabular}  
	\end{threeparttable} 
\end{table*}

\subsection{Initializing Voxel-Point Neuron}
\label{sec:ini_neuron}

The initializing Voxel-Point neuron is used to initialize the feature information from the input 3D data $\textit{\textbf{X}} = \{\textit{\textbf{x}}_1,...,\textit{\textbf{x}}_n\} =\{(\textit{\textbf{p}}_1, \textit{\textbf{f}}_1),...,(\textit{\textbf{p}}_n, \textit{\textbf{f}}_n)\} \subseteq \mathbb{R}^{3+C_{1}}$, where $\textit{\textbf{p}}_i$ is the 3D point coordinates $\textit{\textbf{p}}_i = (x_i, y_i, z_i)$, it may also contain further information, e.g. RGB color and normal; $\textit{\textbf{f}}_i$ is the output feature of previous layer which includes $C_{1}$ channels. 

\subsubsection{Voxel-based Module of Initializing Neuron}
\label{sec:voxel_module}

3D CNN on voxel grid is a popular selection for state-of-the-art 3D deep learning researches. Due to its high regularity and efficient structuring with 3D CNN, we adopt the voxel-based module to capture the initializing neighboring information for Voxel-Point neuron. 

\noindent \textbf{Point Transformation.} Before we implement the Voxel-based module, we conduct point transformation to eliminate the influence from the translation and the scale variations of 3D points. Firstly, we calculate the mean point $\bar{\textit{\textbf{p}}}$ of the input 3D data, and translate each point with $\bar{\textit{\textbf{p}}}$ by $\textit{\textbf{p}}_{i} = \textit{\textbf{p}}_{i} - \bar{\textit{\textbf{p}}}$. Then we search the farthest point $\|\textit{\textbf{p}}\|_{max}$ as the radius, and transform the whole points into the unit sphere:
\begin{equation}
	\hat{\textit{\textbf{p}}}_{i} = \dfrac{\textit{\textbf{p}}_{i}}{2*\|\textit{\textbf{p}}\|_{max}} + 0.5
\end{equation}
where $\hat{\textit{\textbf{p}}}_{i}$ is the transformed point coordinates. During the experiments, we find that the transformation for feature is not necessary, as it reduces the accuracy results of our model.
So, among this process, we just transform the point coordinates, and here we marked the transformed data as $\{\hat{\textit{\textbf{p}}}_{i}, \hat{\textit{\textbf{f}}}_{i}\}$.

\noindent \textbf{3D Voxel CNN.} After the point transformation, the coordinate range of each point is from 0 to 1, thus we enlarge the coordinate value from 0 to $r-1$, where $r$ is the resolution of voxel grid, and we marked the enlarged coordinates as $\hat{\textit{\textbf{p}}}_{ri} (\hat{x}_{ri},\hat{y}_{ri},\hat{z}_{ri})$. Then the points $\hat{\textit{\textbf{p}}}_{ri}$ and its corresponding features $\hat{\textit{\textbf{f}}}_{i}$ are voxelized into the voxels with low spatial resolution of $r \times r \times r$, and the features of voxels are the mean features of all inside points:
\begin{equation}
	\textit{\textbf{F}}(u,v,w) = \frac{\sum_{i=1}^{n} floor (\hat{x}_{ri},\hat{y}_{ri},\hat{z}_{ri}) \cdot \hat{\textit{\textbf{f}}}_{i}}{n}
\end{equation}
where $\textit{\textbf{F}}(u,v,w)$ is the features of voxels $(u,v,w)$, $n$ is inside point number of $(u,v,w)$, and $floor$ is the down rounding function.
The influence of voxel resolution $r$ will be discussed in Section~\ref{sec:ablation_study}.
Next a series of $3 \times 3 \times 3$ 3D CNNs are adopted to aggregate the neighboring feature information of voxels. To capture the neighborhood information more accurately, we have increased the feature channels to $C_2$ during 3D CNN implementation.

\noindent \textbf{Voxel Feature Interpolation.} Then we interpolate the voxel features into the common domain of point cloud since we need to aggregate the information of voxel-based module and point-based module. The common operation for interpolation is the tri-linear interpolation and nearest-neighbor interpolation. Similar to the conduction in PVCNN, we adopt tri-linear interpolation to transform the voxel features into the point features $\textit{\textbf{V}}_{1} = \{V_{11},...,V_{1n}\} \subseteq \mathbb{R}^{C_{2}}$.


\subsubsection{Point-based Module of Initializing Neuron}
\label{sec:point_module}

Although the voxel-based module aggregates the neighboring feature information for the input 3D data, its extracted information is in a coarse manner with low voxel resolution. In order to make up for this defect, we adopt point-based SharedMLP (1D convolution with kernel of 1) to make either efficient learning on each point for the input 3D data. For the sake of convenient aggregation of voxel-based module and point-based module, the output feature channels of point-based module are consistent with the output channels of voxel-based module. Without losing generality, the output features are implemented with batch normalization~\cite{ioffe2015batch} and ReLU activation function~\cite{glorot2011deep}, and these point-based features are marked as $\textit{\textbf{P}}_{1} = \{P_{11},...,P_{1n}\} \subseteq \mathbb{R}^{C_{2}}$. 

When both the feature information of voxel-based and point-based module in is obtained, the  initializing Voxel-Point neuron will output $\textit{\textbf{V}}_{1}$ and $\textit{\textbf{P}}_{1}$ to the transmission Voxel-Point neuron to strengthen the point-based and voxel-based features. 

\begin{figure*}[t]
	\begin{center}
		\includegraphics[width=0.95\linewidth]{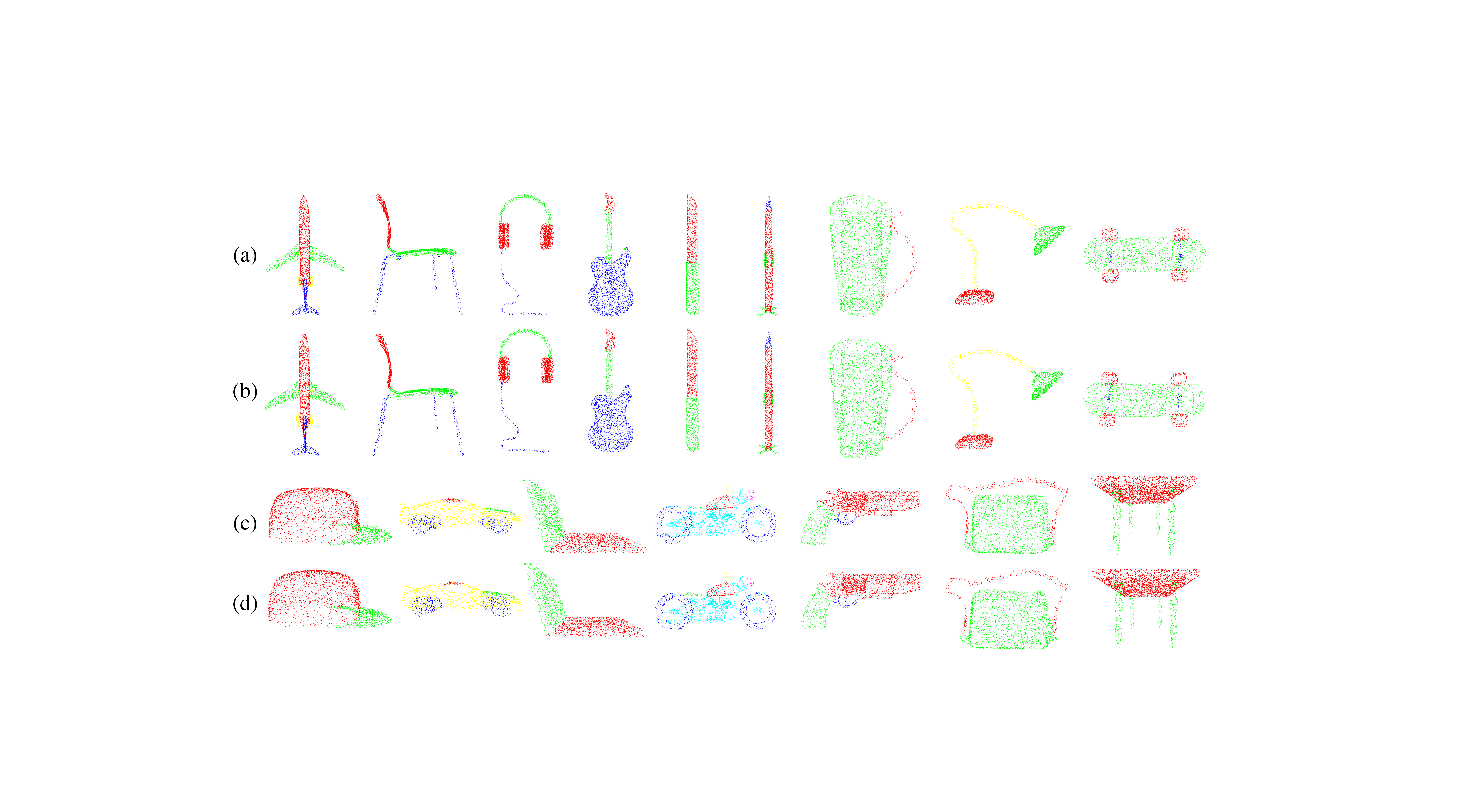}
	\end{center}
	\caption{Segmentation results on ShapeNet Part. (a)(c) Ground truth. (b)(d) Our results.}
	\label{fig:vis_shapenet}
\end{figure*}

\subsection{Transmission Voxel-Point Neuron}
\label{sec:trans_neuron}

The transmission Voxel-Point neuron is used to strengthen the fusion features from the initializing Voxel-Point neuron. The enhancement of feature information is reflected in two aspects: (1) increasing the neighboring collection for the individual point-based features $\textit{\textbf{P}}_{1}$; (2) increasing the independence of the neighboring voxel-based features $\textit{\textbf{V}}_{1}$.

\noindent \textbf{Neighboring collection for individual point-based feature.} Transmission Voxel-Point neuron obtains the input features from the output aggregation feature information of initializing Voxel-Point neuron. In terms of network structures, the point-based module in transmission Voxel-point neuron is the same as the one in initializing Voxel-Point neuron, the voxel-based module got a little differences. The voxel-based and point-based information from the last Voxel-Point neuron have aggregated together before inputting to the next Voxel-Point neuron, but as we know, the voxel-based features $\textit{\textbf{V}}_{1}$ have carried out information exchanging between different channels with 3D CNN, and yet the point-based features $\textit{\textbf{P}}_{1}$ haven't exchanged. 
To enhance the neighboring collection for the point-based features $\textit{\textbf{P}}_{1}$ from the last neuron with 3D CNN in the voxel-based module of the current transmission Voxel-Point neuron, we firstly fuse $\textit{\textbf{P}}_{1}$ and $\textit{\textbf{V}}_{1}$ to obtain the fused features $\{\textit{\textbf{V}}_{1}+\textit{\textbf{P}}_{1}\}\subseteq \mathbb{R}^{C_{2}}$, then voxelize $\{\textit{\textbf{V}}_{1}+\textit{\textbf{P}}_{1}\}$ with low spatial resolution $r \times r \times r$. In order to further strengthen the neighboring collection for $\textit{\textbf{P}}_{1}$ and exchange the collection between different channels of $\textit{\textbf{V}}_{1}$, we set a 3D CNN with kernel of $1 \times 1 \times 1$ in the voxel-based module of the current transmission Voxel-Point neuron before the regular $3 \times 3 \times 3$ 3D CNN operations, thus we obtain the enhanced features $\textit{\textbf{V}}_{2} = \{V_{21},...,V_{2n}\} \subseteq \mathbb{R}^{C_{2}}$.

\noindent \textbf{Independence for the neighboring voxel-based features.} To add the independent attribute for voxel-based features $\textit{\textbf{V}}_{1}$, and increase the fine granularity of point-based features $\textit{\textbf{P}}_{1}$ by the way, we adopt shared MLP to the fused features $\{\textit{\textbf{V}}_{1}+\textit{\textbf{P}}_{1}\}\subseteq \mathbb{R}^{C_{2}}$. The parameters and network structure are consistent with the point-based module of initializing Voxel-Point neuron, and here we obtain the enhanced individual features $\textit{\textbf{P}}_{2} = \{P_{21},...,P_{2n}\} \subseteq \mathbb{R}^{C_{2}}$.
After the completion of point-based module and voxel-based module in the transmission Voxel-Point neuron, next is to output the aggregating information. 
To output the final features, in addition to the transmission neuron's aggregation information, the initializing neuron's voxel-based features are also aggregated to the output, and thus we output $\{\textit{\textbf{V}}_{1}+\textit{\textbf{V}}_{2}+\textit{\textbf{P}}_{2}\}\subseteq \mathbb{R}^{C_{2}}$ as the final features. In Section~\ref{sec:ablation_study}, we'll discuss the output aggregating features. 

\section{Experimental Results}
\label{sec:exp}

\renewcommand{\arraystretch}{1.0}  
\renewcommand\tabcolsep{3.0pt}
\begin{table*}[ht]
	\centering  
	\fontsize{9}{11}\selectfont  
	\begin{threeparttable}  
		\caption{S3DIS}  
		\label{tab:s3dis}  
		\begin{tabular}{lccccccccccccccccccccccc}  
			\toprule  
			Method&Reference&Type&Input Data&mIoU&mAcc&GPU Mem.&latency\cr  
			\midrule
			mIoU $<$ 56.0\cr
			PointNet~\cite{qi2017pointnet}&CVPR 2017&point-based&$8 \times 4K$ points&42.97&82.54&0.6GB&\textbf{20.9ms}\cr
			DGCNN~\cite{wang2019dynamic}&SIGGRAPH 2019&point-based&$8 \times 4K$ points&47.94&83.64&2.4GB&178.1ms\cr
			3D-GCN~\cite{lin2021learning}&TPAMI 2021&point-based&$8 \times 4K$ points&51.90&84.60&-&-\cr
			RSNet~\cite{huang2018recurrent}&CVPR 2018&point-based&$8 \times 4K$ points&51.93&-&1.1GB&111.5ms\cr
			PointNet++~\cite{qi2017pointnetplusplus}&NeurIPS 2017&point-based&$8 \times 4K$ points&52.28&-&-&-\cr
			TanConv~\cite{tatarchenko2018tangent}&CVPR 2018&point-based&$8 \times 4K$ points&52.8&85.5&-&-\cr
			3D-UNet~\cite{cciccek20163d}&MICCAI 2016&voxel-based&$8 \times 96^3$ voxels&54.93&86.12&6.8GB&574.7ms\cr
			JSNet~\cite{zhao2020jsnet}&AAAI 2020&point-based&$8 \times 4K$ points&54.5&87.7&-&-\cr
			SSNet~\cite{thabet2020self}&CVPR 2020&point-based&-&55.00&61.20&-&-\cr
			$\textbf{MVPCNN}_{(0.25 \times Ch)}$&-&voxel-point-based&$8 \times 4K$ points&\textbf{55.30}&\textbf{86.91}&2.3GB&51.9ms\cr 
			\midrule
			mIoU $<$ 58.9\cr	
			PVCNN~\cite{liu2019pvcnn}&NeurIPS 2019&voxel-point-based&$8 \times 4K$ points&56.12&86.66&1.3GB&47.3ms\cr
			PointCNN~\cite{Li2019pointcnn}&NeurIPS 2018&point-based&$16 \times 2K$ points&57.26&85.91&4.6GB&282.3ms\cr
			Grid-GCN~\cite{xu2020grid}&CVPR 2020&point-based&$8 \times 4K$ points&57.75&86.94&-&\textbf{25.9ms}\cr
			$\textbf{MVPCNN}_{(1 \times Ch)}$&-&voxel-point-based&$8 \times 4K$ points&\textbf{58.63}&\textbf{87.75}&4.3GB&72.7ms\cr
			\midrule
			mIoU $>$ 58.9\cr
			SPNet~\cite{liu2020self}&ECCV 2020&point-based&$8 \times 4K$ points&58.80&65.90&-&-\cr
			CF-SIS~\cite{wen2020cf}&ACM MM 2020&point-based&$8 \times 4K$ points&58.90&67.30&-&-\cr
			PVCNN++~\cite{liu2019pvcnn}&NeurIPS 2019&voxel-point-based&$4 \times 8K$ points&58.98&87.12&0.8GB&69.5ms\cr
			$\textbf{MVPCNN++}_{(0.5 \times Ch)}$&-&voxel-point-based&$4 \times 8K$ points&60.17&88.76&2.7GB&\textbf{52.6ms}\cr
			$\textbf{MVPCNN++}_{(1 \times Ch)}$&-&voxel-point-based&$4 \times 8K$ points&\textbf{61.51}&\textbf{89.31}&4.3GB&72.7ms\cr
			\bottomrule  
		\end{tabular}  
	\end{threeparttable} 
\end{table*}  

In this section, we firstly introduce the implementation details of MVPConv, then we compare our method with the state-of-the-art frameworks on various 3D datasets for different tasks, e.g. ShapeNet Parts (object part segmentation)~\cite{chang2015shapenet}, S3DIS (indoor scene segmentation)~\cite{armeni20163d,armeni2017joint} and KITTI (3D object detection)~\cite{geiger2013vision}. Finally, we carry on additional ablation study to certificate our proposed idea.

\subsection{Implementation Details}
\label{sec:mvp_detail}

As is shown in Fiure~\ref{fig:mvpconv}, MVPConv consists of initializing and transmission neurons, both of these two neurons including voxel-based and point-based modules, the whole framework is implemented by using Pytorch. 

\textbf{The initializing Voxel-Point neuron}, its voxel-based modules contain two $3 \times 3 \times 3$ 3D CNNs with stride 1 and padding 1, each 3D CNN is followed by 3D batch normalization~\cite{ioffe2015batch} and Leaky ReLU activation function~\cite{maas2013rectifier}. The point-based SharedMLP is a 1D CNN with kernel of 1 which converts the feature channels to make it consistent with the output feature channels of voxel-based module, next to the 1D CNN is 1D batch normalization~\cite{ioffe2015batch} and ReLU activation function~\cite{glorot2011deep}.

\textbf{The transmission Voxel-Point neuron} is based on the initializing voxel-point neuron, it sets the output aggregation-fused feature of the initializing voxel-point neuron as input. Its voxel-based modules contain one $1 \times 1 \times 1$ 3D CNN with stride 1 and padding 0, and the rest part is the same as the one in the voxel-based modules of the initializing voxel-point neuron. The point-based module also keeps the output feature channels consistent with its voxel-based module.

\begin{figure*}[t]
	\begin{center}
		\includegraphics[width=0.95\linewidth]{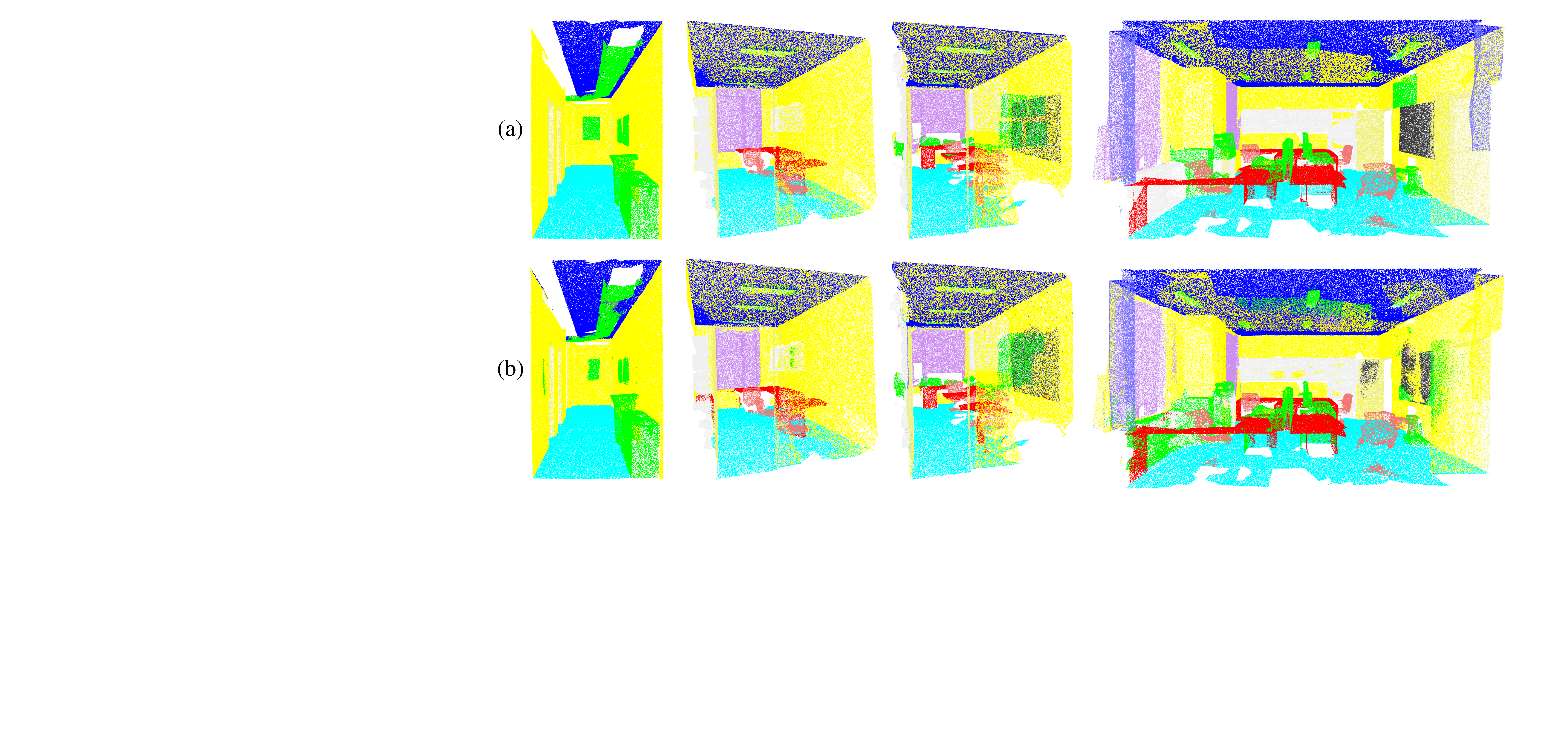}
	\end{center}
	\caption{Segmentation results on S3DIS area5. (a) Ground truth. (b) Our results. }
	\label{fig:vis_s3dis}
\end{figure*}

\subsection{Part Segmentation}
\label{sec:shapenet}

\textbf{Dataset.} We extend our MVPConv for part segmentation on ShapeNet part dataset~\cite{chang2015shapenet}. ShapeNet part dataset including 16881 3D shapes of 16 object categories, and it is with 50 parts in all. We sample 2048 points from each shape as input training data. To be fair, we follow the typical evaluation shceme as Liu \etal~\cite{liu2019pvcnn}, Li \etal~\cite{Li2019pointcnn} and Graham \etal~\cite{graham20183d} in our experiments. 

\textbf{Architecture.} We use the PointNet as the backbone, and our MVPCNN is built by updating the shared MLP layers in PointNet~\cite{qi2017pointnet} with our proposed MVPConv layers. 

\textbf{Training.} We set the batch size to 8, and adopt the ADAM optimizer~\cite{kingma2014adam} with learning rate of 0.001 for 200 epochs. The criterion is adopted with CrossEntropy. The whole training process is carried out on a single RTX 2080Ti GPU.

\textbf{Results.} The mean intersection-over-union (mIoU) is utilized as evaluation metric. We adopt the same evaluation scheme in PointNet, where the IoU of each shape is calculated by averaging the same shape parts' IoUs of the 2874 test models. The mIoU is obtained by averaging the IoU of all shapes. We compare our model against the state-of-the-art point-based methods~\cite{klokov2017escape,qi2017pointnet,li2018so,xie2018attentional,su2018splatnet,shen2018mining,huang2018recurrent,qi2017pointnetplusplus,wang2019dynamic,atzmon2018point,xu2018spidercnn,xu2020geometry,lin2021learning,Li2019pointcnn,liu2020self,wen2020cf,liu2019relation}, voxel-based methods~\cite{cciccek20163d} and the newest point-voxel-based model~\cite{liu2019pvcnn}.
To better balance the trade-off between time efficiency and accuracy, we also reduce the output feature channels to 50\% and 25\%, and marked as $\textbf{MVPCNN}_{(0.5 \times Ch)}$ and $\textbf{MVPCNN}_{(0.25 \times Ch)}$ respectively. Table~\ref{tab:shapenet} and Figure~\ref{fig:vis_shapenet} presents the evaluation results of MVPCNN on the ShapeNet part dataset. Our MVPCNN outperforms the PointNet backbone significantly with 3.3\% increase of mIoU, even if we reduce the latency by 8.8\%, we can still obtain 2.2\% mIoU ahead of PointNet. Compare with the voxel-based 3D-UNet and point-based SpiderCNN, we are $\textbf{34} \times$ faster and $\textbf{8} \times$ respectively, while achieving the better accuracy. In addition, we also outperform other point-based methods with better accuracy and higher efficiency. Notably, even compared with the voxel-point-based PVCNN, we still gain higher mIoU by 0.3\% with a small amount of time efficiency lost, and PVCNN is a state-of-the-art method aimed at speed improvement.

\subsection{Indoor Scene Segmentation}
\label{sec:s3dis}

\textbf{Dataset.} Stanford large-scale 3D Indoor Spaces (S3DIS) Dataset~\cite{armeni20163d,armeni2017joint} is a benchmark used for semantic indoor scene segmentation. S3DIS consists of 6 3D scanning indoor areas which totally includes 272 rooms. Similar to the operation in Liu \etal~\cite{liu2019pvcnn}, Li \etal~\cite{Li2019pointcnn} and Tchapmi \etal~\cite{tchapmi2017segcloud}, we set Area-1, 2, 3, 4, 6 as the training set, and the rest Area-5 is used to test as it's the only one that has no overlapping with others. We follow the data preprocessing and the evaluation criteria as Li \etal~\cite{Li2019pointcnn} before training, each block is sampled to 4096 points for training. 

\textbf{Architecture.} In this task we also use PointNet as the baseline (MVPCNN), moreover, we also adopt PointNet++ as the backbone to build MVPCNN++ with MVPCNN. Similar to the experiments on ShapeNet part dataset in Section~\ref{sec:shapenet}, we also design a compressed version by reducing the output feature channels to 12.5\%, 25\% and 50\% in MVPCNN, and 50\% in MVPCNN++.

\textbf{Training.} The batch size, optimizer, learning rate and criterion are set the same as what we have done in ShapeNet part experiments. The number of epoch is set to 50. It's also implemented in a single RTX 2080Ti GPU.

\textbf{Results.} Apart from mIoU, mean accuracy (mAcc) is also used to evaluate the performance of our proposed model. In addition to comparing with the state-of-the-art point-based~\cite{qi2017pointnet,wang2019dynamic,lin2021learning,huang2018recurrent,qi2017pointnetplusplus,tatarchenko2018tangent,zhao2020jsnet,thabet2020self,Li2019pointcnn,xu2020grid,liu2020self,wen2020cf} and voxel-based methods~\cite{cciccek20163d}, we also compare with the newest point-voxel-based model~\cite{liu2019pvcnn}. 
Table~\ref{tab:s3dis} shows the results of all methods on S3DIS dataset, and Figure~\ref{fig:vis_s3dis} presents the visualization results. Our MVPCNN improves the mIoU of the backbone (PointNet) by more than \textbf{36\%}, and MVPCNN++ outperforms its backbone (PointNet++) by a large margin in mIoU of more than \textbf{17\%}. Notably, The compact MVPCNN of 25\% feature channel outperforms the voxel-based 3D-UNet in accuracy with more than \textbf{11}$\times$ lower latency, and the point-based DGCNN by more than \textbf{15\%} in mIoU with $\textbf{3} \times$ lower latency. Significantly, the full model of MVPCNN outperforms the state-of-the-art point-based model (Grid-GCN and PointCNN), and we increase the speed by nearly \textbf{4}$\times$ compared with PointCNN. At the same time, MVPCNN also outperforms the newest voxel-point-based method (PVCNN) both in mIoU and mAcc. Remarkably, the compact version of MVPCNN++ is faster than the extremely efficient voxel-point-based PVCNN++, and also outperforms it in accuracy.The full MVPCNN++ can improve the performance more than \textbf{4}\% in mIoU compared with PVCNN++.

\subsection{3D Detection}
\label{sec:kitti}

\renewcommand{\arraystretch}{1.2}  
\begin{table*}[ht]  
	\centering  
	\fontsize{9}{10}\selectfont  
	\begin{threeparttable}  
		\caption{KITTI}  
		\label{tab:kitti}  
		\begin{tabular}{lcccccccccccc}  
			\toprule  
			\multicolumn{1}{c}{}&  
			\multicolumn{3}{c}{Car}&\multicolumn{3}{c}{Pedestrian}&\multicolumn{3}{c}{Cyclist}&\multicolumn{2}{c}{Efficiency}\cr  
			\cmidrule(lr){2-4} \cmidrule(lr){5-7}  \cmidrule(lr){8-10} \cmidrule(lr){11-12}
			&Easy &Mod.&Hard&Easy&Mod.&Hard&Easy&Mod.&Hard&GPU Mem.&latency\cr  
			\midrule  
			F-PointNet~\cite{qi2018frustum}&85.24&71.63&63.79&66.44&56.90&50.43&77.14&56.46&52.79&1.3GB&29.1ms\cr  
			F-PointNet++~\cite{qi2018frustum}&84.72&71.99&64.20&68.40&60.03&52.61&75.56&56.74&53.33&2.0GB&105.2ms\cr  
			F-PVCNN~\cite{liu2019pvcnn}&85.25&72.12&64.24&70.60&61.24&56.25&78.10&57.45&53.65&1.4GB&58.9ms\cr  
			F-MVPCNN&\textbf{85.66}&\textbf{72.63}&\textbf{64.62}&\textbf{71.12}&\textbf{62.34}&\textbf{57.13}&\textbf{79.85}&\textbf{58.28}&\textbf{54.62}&2.2GB&100.0ms\cr
			\bottomrule  
		\end{tabular}  
	\end{threeparttable}  
\end{table*}

\textbf{Dataset.} We evaluate our model on KITTI~\cite{geiger2013vision} for 3D detection task. This dataset is a benchmark for autonomous driving which contains 7481 training and 7518 test samples. For fair comparison, we follow the treatment processes as Qi \etal~\cite{qi2018frustum} in our tests. Among the 7481 training samples, there are 3712 samples splitting as the \textit{train} set, the rest 3769 is set as \textit{val} set.

\textbf{Architecture.} Without changing the whole network, we set F-PointNet~\cite{qi2018frustum} as the backbone by replacing the shared MLP layers of the instance segmentation network with our MVPConv to generate F-MVPCNN. 

\textbf{Training.} The same setting for training as in part segmentation and indoor scene segmentation except that epoch is set to 209 and batch size is set to 32.

\textbf{Results.} Mean average precision (mAP) is adopted to evaluate our model. We compare our model againt F-PointNet, F-PointNet++ and F-PVCNN (whose backbone is also F-PointNet). Tabel~\ref{tab:kitti} shows the experimental results on KITTI. Our F-MVPCNN outperforms all methods in all classes, and ours improves the mAP of F-PointNet (backbone) by up to \textbf{13.2\%}. Compared with F-PointNet++, we outperform the performances of it with faster speed, especially in the hard pedestrian class.

\subsection{Ablation Study}
\label{sec:ablation_study}

In this part, the additional ablation experiments are conducted to analyze the design idea of MVPConv. All frameworks are trained and tested on ShapeNet part datasets~\cite{chang2015shapenet}, and we consider the half of the feature channels to output as it is a good trade-off between accuracy and latency.  

\textbf{Effects of transmission Voxel-Point neuron for MVPConv.} We have tried to enlarge the voxel resolution and directly deepen 3D CNN network of voxel-based module in initializing Voxel-Point neuron before we think about constructing the transmission Voxel-Point neuron for the whole MVPConv.
As presented in the $2_{nd}$ rows of Table~\ref{tab:cnn_res}, the increasing of voxel resolution could improve the performance of network in segmentation task a little bit, but the GPU memory and latency increase dramatically. Conversely, the $3_{rd}$ rows of Table~\ref{tab:cnn_res} show that the accuracy of deepening network structure by 3D convolution is reduced on the contrary, and the GPU memory and latency increased also, but the increase of them almost can be neglected.
So we tried another way to deepen the network, that is the transmission Voxel-Point neuron. As shown in the $4_{th}$ rows of Table~\ref{tab:cnn_res}, the transmission Voxel-Point neuron improves significant performance of MVPConv without consuming much more GPU memory and latency. 

\renewcommand{\arraystretch}{1.0}  
\renewcommand\tabcolsep{3.0pt}
\begin{table}[ht]
	\centering  
	\fontsize{9}{11}\selectfont  
	\begin{threeparttable}  
		\caption{Effects of transmission Voxel-Point neuron.}  
		\label{tab:cnn_res}  
		\begin{tabular}{lcccccccccccccccccccccc}  
			\toprule  
			Model&mIoU&GPU Mem.&latency\cr  
			\midrule
			Init. Neuron&85.50&1.97GB&22.3ms\cr
			Init. Neuron (1.5$\times$R)&85.55&2.35GB&37.1ms\cr
			Init. Neuron (3$\times$Conv3D)&85.33&2.08GB&25.1ms\cr
			Init. Neuron + Tran. Neuron&\textbf{85.76}&2.11GB&31.0ms\cr
			\bottomrule  
		\end{tabular}  
	\end{threeparttable} 
\end{table}  

\textbf{Effects of different features for MVPConv.} As shown in Table~\ref{tab:feature_selection}, we study the importance of the four feature components of MVPConv. The $1_{st}$ row indicates that the performance of MVPConv is greatly degraded if only the feature of $\textbf{\textit{V}}_{2}$ is aggregated, as the voxel-based feature $\textbf{\textit{V}}_{2}$ is not enough for 3D learning. The feature aggregations of D, F, G and H significantly improve the performances of MVPConv, as $\textbf{\textit{V}}_{2}$, and the combination of $\textbf{\textit{V}}_{1}+\textbf{\textit{V}}_{2}+\textbf{\textit{P}}_{2}$ achieves the best performance.

\renewcommand{\arraystretch}{1.0}  
\renewcommand\tabcolsep{3.0pt}
\begin{table}[ht]
	\centering  
	\fontsize{9}{11}\selectfont  
	\begin{threeparttable}  
		\caption{Effects of different features for MVPConv.}  
		\label{tab:feature_selection}  
		\begin{tabular}{cccccccccccccccccccccccc}  
			\toprule  
			Model\quad&\quad$\textbf{\textit{V}}_{1}$\quad&$\quad\textbf{\textit{P}}_{1}$\quad&\quad$\textbf{\textit{V}}_{2}$\quad&\quad$\textbf{\textit{P}}_{2}$\quad&\quad mIoU\cr  
			\midrule
			A\quad&\quad \quad&\quad \quad&\quad \checkmark \quad&\quad \quad&\quad 85.39\cr
			B\quad&\quad \checkmark \quad&\quad \checkmark \quad&\quad \quad&\quad \quad&\quad 85.50\cr
			C\quad&\quad \quad&\quad \checkmark \quad&\quad \checkmark \quad&\quad \quad&\quad 85.43\cr
			D\quad&\quad \quad&\quad \quad&\quad \checkmark \quad&\quad \checkmark \quad&\quad 85.58\cr
			E\quad&\quad \checkmark \quad&\quad \checkmark \quad&\quad \checkmark \quad&\quad \quad&\quad85.46\cr
			F\quad&\quad \quad&\quad \checkmark \quad&\quad \checkmark \quad&\quad \checkmark \quad&\quad85.54\cr
			G\quad&\quad \checkmark \quad&\quad \quad&\quad \checkmark \quad&\quad \checkmark \quad&\quad \textbf{85.76}\cr
			H\quad&\quad \checkmark \quad&\quad \checkmark \quad&\quad \checkmark \quad&\quad \checkmark \quad&\quad85.57\cr	
			\bottomrule  
		\end{tabular}  
	\end{threeparttable} 
\end{table}

\textbf{Effects of 1}$\times$\textbf{1}$\times$\textbf{1 3D CNN for MVPConv.} In Table~\ref{tab:1conv}, we research the effect of 1$\times$1$\times$1 3D CNN for MVPConv. 1$\times$1$\times$1 3D CNN can strengthen the information association and nonlinearity between different feature channels of point-based and voxel-based features, thus could improve the 3D learning ability of the network. As presented, almost no increase in latency and GPU memory, 1$\times$1$\times$1 3D CNN helps MVPConv achieve better results of accuracy. 

\renewcommand{\arraystretch}{1.0}  
\renewcommand\tabcolsep{3.0pt}
\begin{table}[ht]
	\centering  
	\fontsize{9}{11}\selectfont  
	\begin{threeparttable}  
		\caption{Effects of 1$\times$1$\times$1 3D CNN for MVPConv.}  
		\label{tab:1conv}  
		\begin{tabular}{lcccccccccccccccccccccc}  
			\toprule  
			Model&mIoU&GPU Mem.&latency\cr  
			\midrule
			MVPConv&85.72&2.024GB&31.2ms\cr
			MVPConv (1$\times$1$\times$1 CNN)&\textbf{85.76}&2.033GB&31.6ms\cr
			\bottomrule  
		\end{tabular}  
	\end{threeparttable} 
\end{table}  



\section{Conclusion}
\label{sec:conclusion}

We have presented MVPConv, a 3D convolution neural network for fast and accurate 3D deep learning. Our method integrates both the voxels and points to construct initializing and transistion Point-voxel neurons, which increases the neighboring collection between point-based features and promote the independence among voxel-based features with efficient convolutions. Experimental results on multiple datasets demonstrate that our proposed method significantly improves the performances of different tasks more efficiently.

{\small
	\bibliographystyle{ieee_fullname}
	\bibliography{egbib}
}

\clearpage
\section*{Supplementary Materials}

\begin{figure*}[t]
	\begin{center}
		\includegraphics[width=0.9\linewidth]{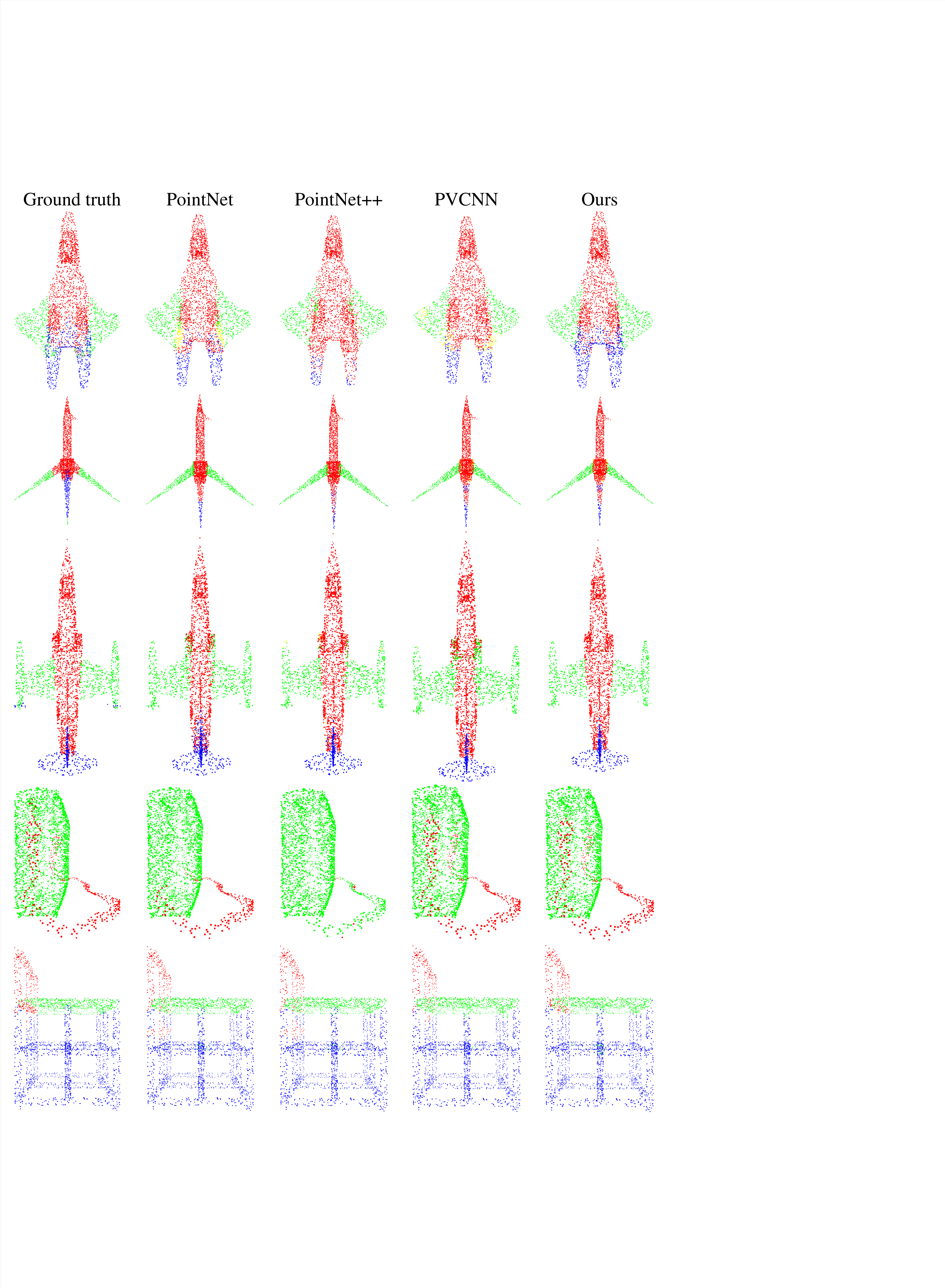}
	\end{center}
	\caption{Part segmentation results of PointNet~\cite{qi2017pointnet}, PointNet~\cite{qi2017pointnetplusplus} PVCNN~\cite{liu2019pvcnn} and our MVPCNN on Shapenet Part.}
	\label{fig:shapenet1}
\end{figure*}

\begin{figure*}[t]
	\begin{center}
		\includegraphics[width=0.8\linewidth]{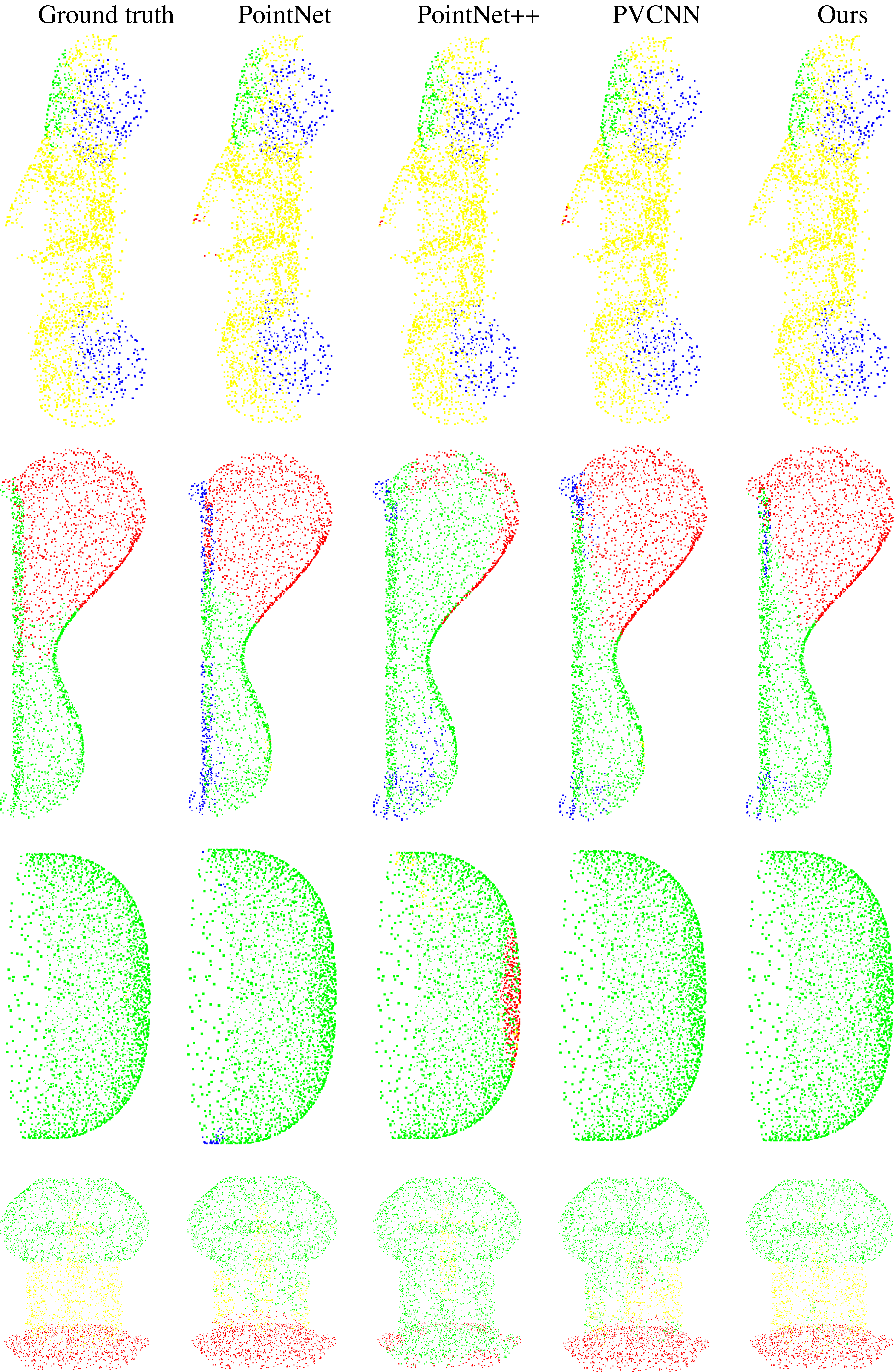}
	\end{center}
	\caption{Part segmentation results of PointNet~\cite{qi2017pointnet}, PointNet~\cite{qi2017pointnetplusplus} PVCNN~\cite{liu2019pvcnn} and our MVPCNN on Shapenet Part.}
	\label{fig:shapenet2}
\end{figure*}

\begin{figure*}[t]
	\begin{center}
		\includegraphics[width=0.85\linewidth]{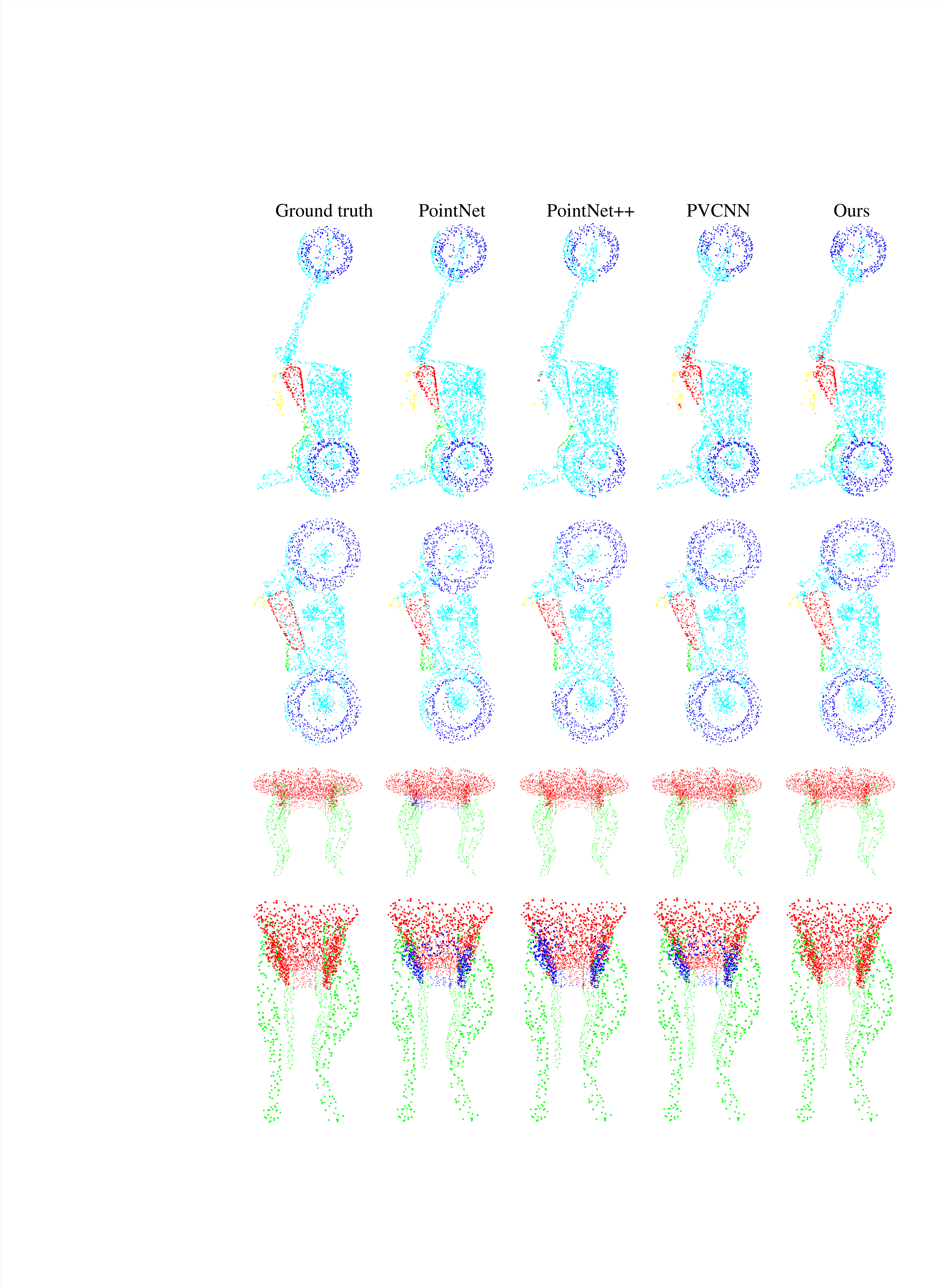}
	\end{center}
	\caption{Part segmentation results of PointNet~\cite{qi2017pointnet}, PointNet~\cite{qi2017pointnetplusplus} PVCNN~\cite{liu2019pvcnn} and our MVPCNN on Shapenet Part.}
	\label{fig:shapenet3}
\end{figure*}

\begin{figure*}[t]
	\begin{center}
		\includegraphics[width=1\linewidth]{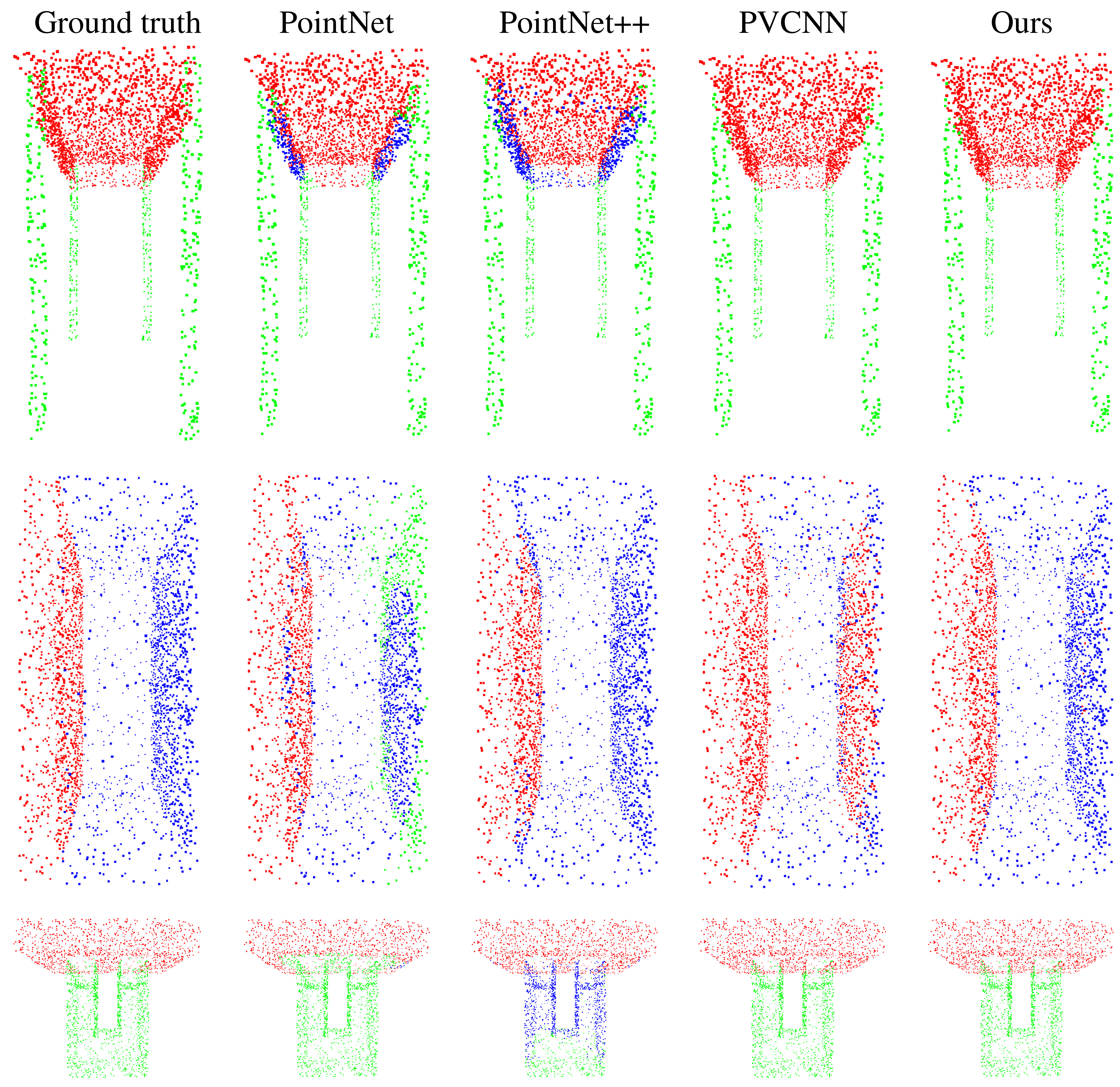}
	\end{center}
	\caption{Part segmentation results of PointNet~\cite{qi2017pointnet}, PointNet~\cite{qi2017pointnetplusplus} PVCNN~\cite{liu2019pvcnn} and our MVPCNN on Shapenet Part.}
	\label{fig:shapenet4}
\end{figure*}

\begin{figure*}[t]
	\begin{center}
		\includegraphics[width=0.81\linewidth]{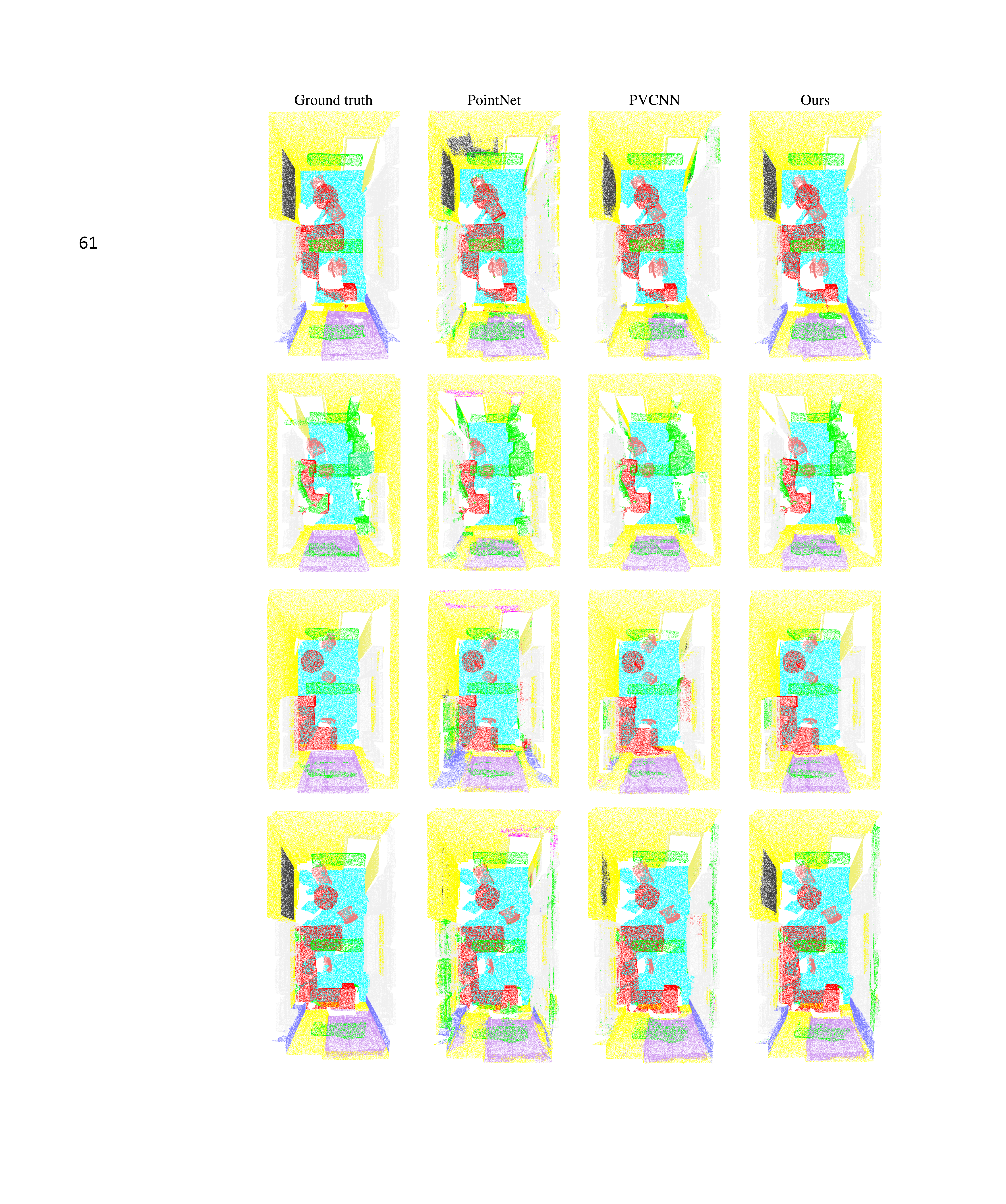}
	\end{center}
	\caption{Indoor scene segmentation results of PointNet~\cite{qi2017pointnet}, PVCNN~\cite{liu2019pvcnn} and our MVPCNN on S3DIS are5.}
	\label{fig:s3dis1}
\end{figure*}

\begin{figure*}[t]
	\begin{center}
		\includegraphics[width=0.93\linewidth]{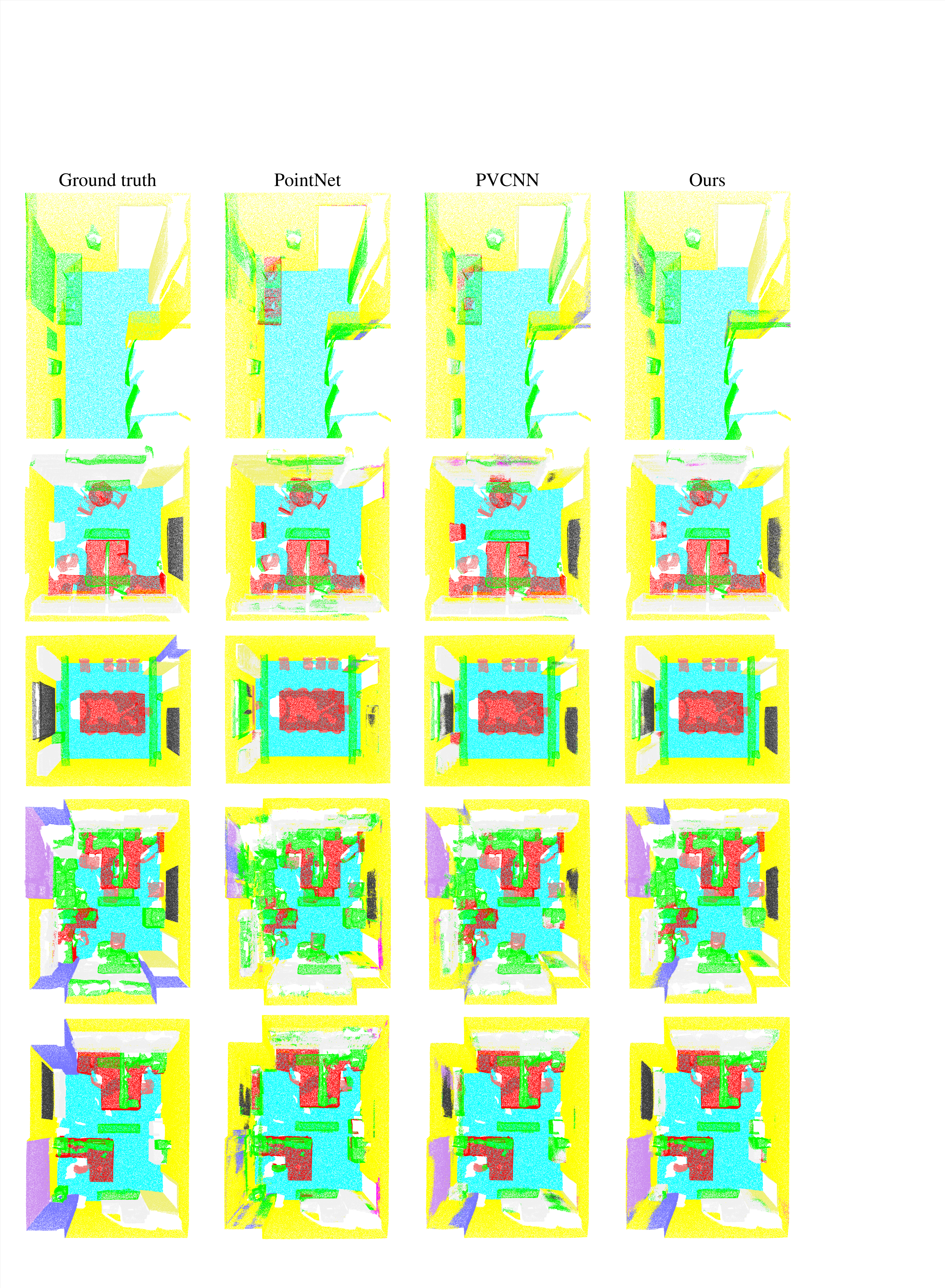}
	\end{center}
	\caption{Indoor scene segmentation results of PointNet~\cite{qi2017pointnet}, PVCNN~\cite{liu2019pvcnn} and our MVPCNN on S3DIS are5.}
	\label{fig:s3dis2}
\end{figure*}

\begin{figure*}[t]
	\begin{center}
		\includegraphics[width=0.78\linewidth]{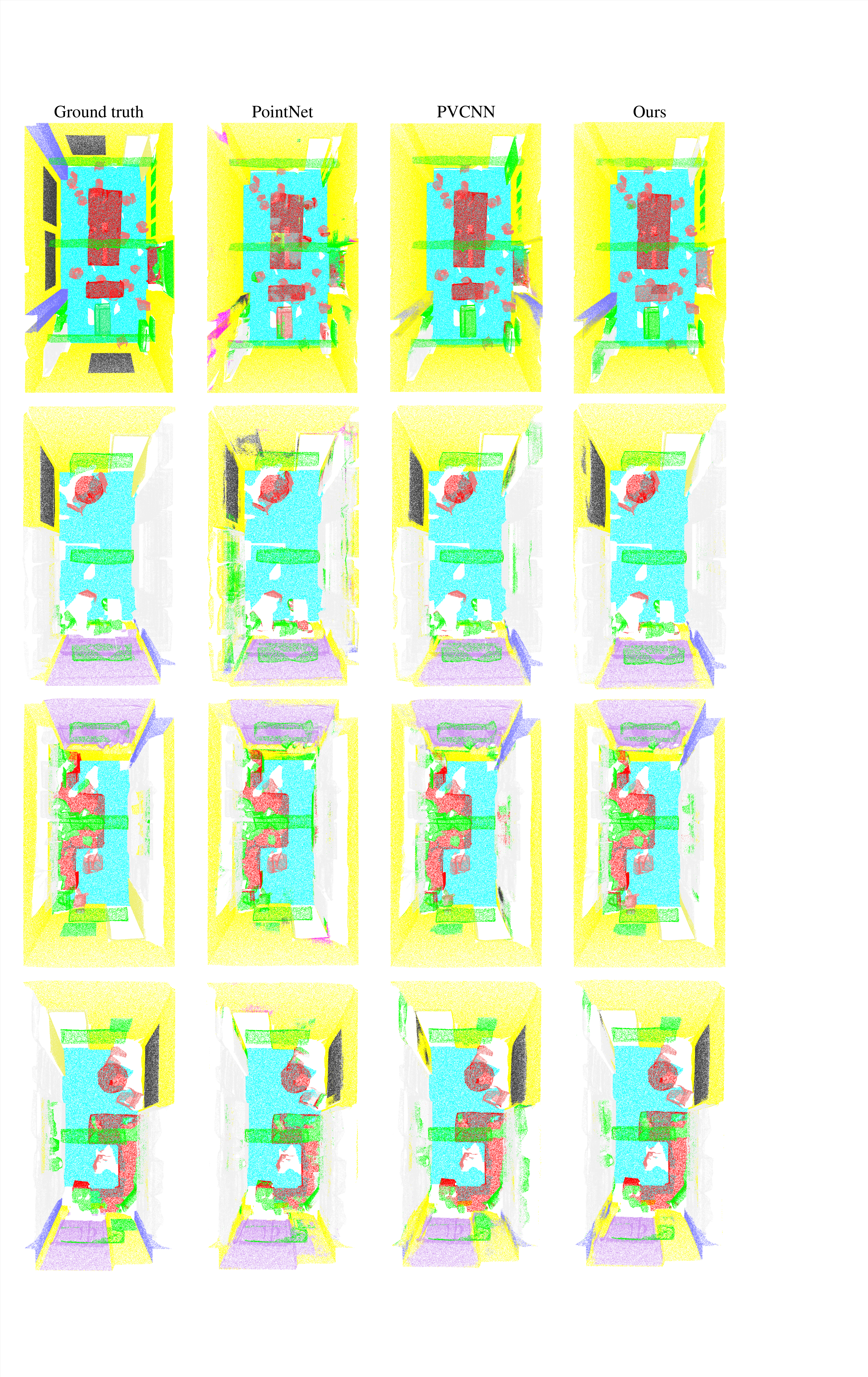}
	\end{center}
	\caption{Indoor scene segmentation results of PointNet~\cite{qi2017pointnet}, PVCNN~\cite{liu2019pvcnn} and our MVPCNN on S3DIS are5.}
	\label{fig:s3dis3}
\end{figure*}

\begin{figure*}[t]
	\begin{center}
		\includegraphics[width=0.77\linewidth]{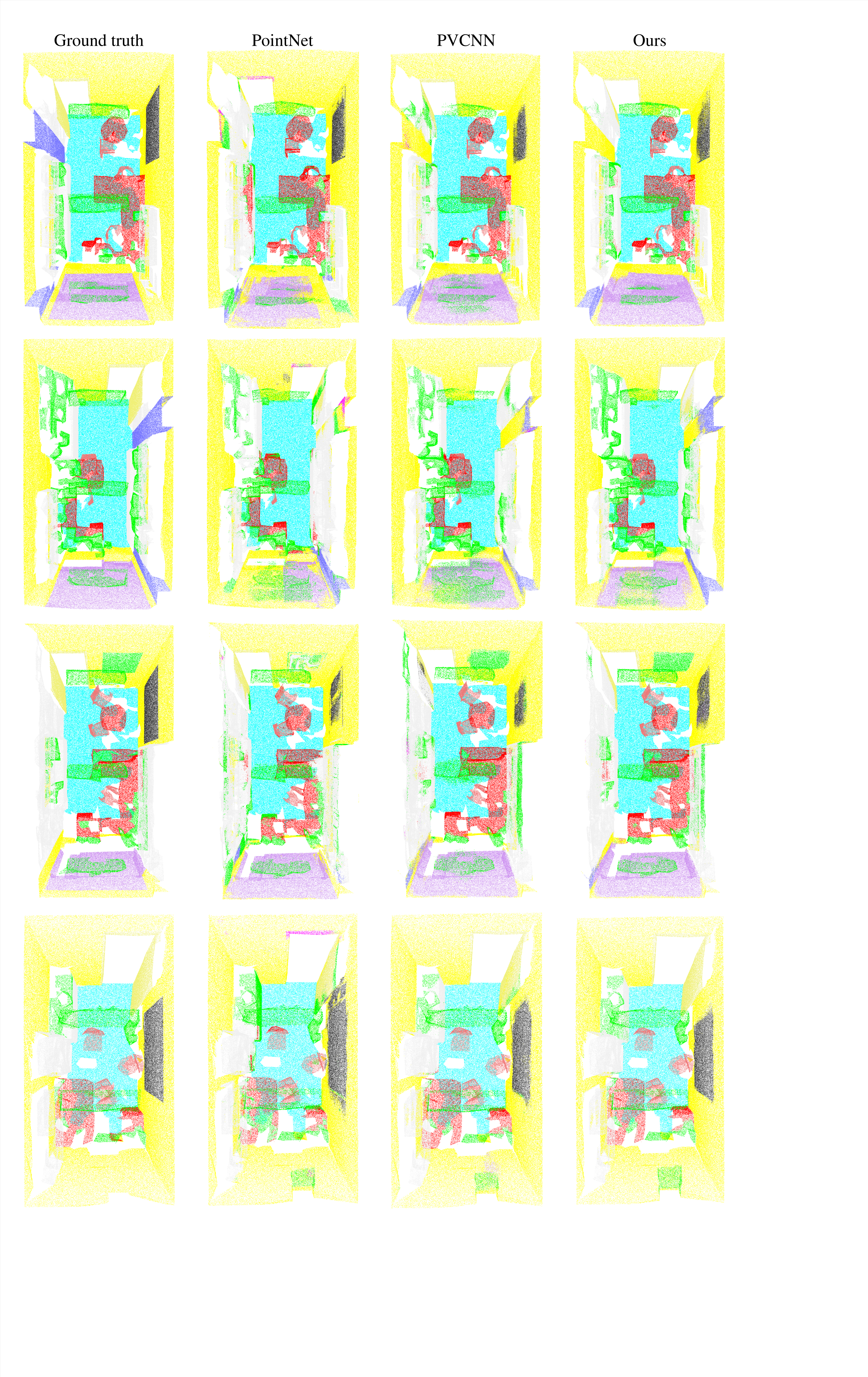}
	\end{center}
	\caption{Indoor scene segmentation results of PointNet~\cite{qi2017pointnet}, PVCNN~\cite{liu2019pvcnn} and our MVPCNN on S3DIS are5.}
	\label{fig:s3dis4}
\end{figure*}

\begin{figure*}[t]
	\begin{center}
		\includegraphics[width=0.82\linewidth]{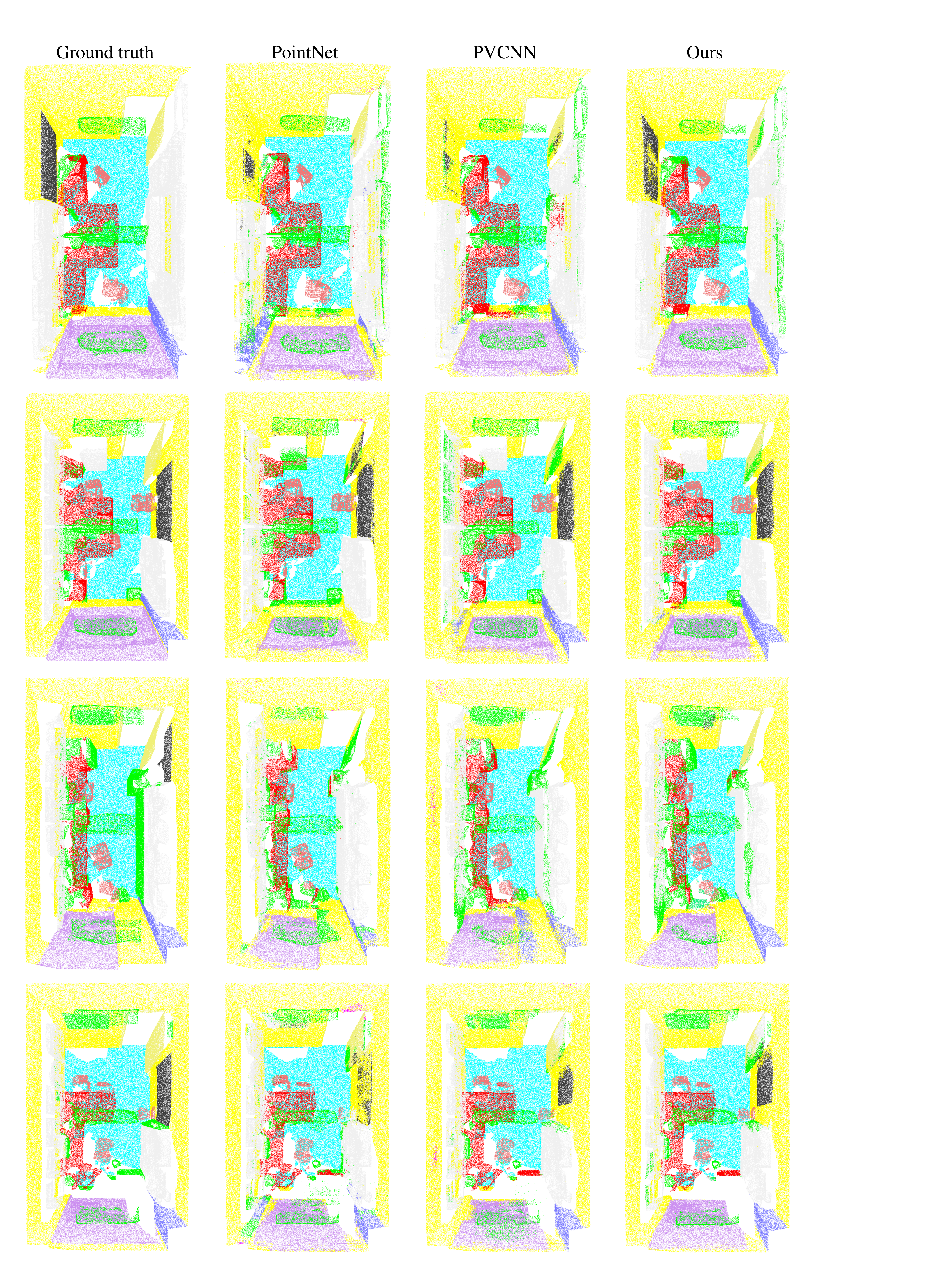}
	\end{center}
	\caption{Indoor scene segmentation results of PointNet~\cite{qi2017pointnet}, PVCNN~\cite{liu2019pvcnn} and our MVPCNN on S3DIS are5.}
	\label{fig:s3dis5}
\end{figure*}

In the supplementary materials, we show more comparable visualization results in part segmentation (Figure~\ref{fig:shapenet1}, Figure~\ref{fig:shapenet2}, Figure~\ref{fig:shapenet3} and Figure~\ref{fig:shapenet4}) and indoor scene segmentation (Figure~\ref{fig:s3dis1}, Figure~\ref{fig:s3dis2}, Figure~\ref{fig:s3dis3}, Figure~\ref{fig:s3dis4} and Figure~\ref{fig:s3dis5}) tasks with PointNet~\cite{qi2017pointnet}, PointNet++~\cite{qi2017pointnetplusplus}, PVCNN~\cite{liu2019pvcnn} and our MVPCNN. 
Besides, we will make our code and models publicly available.

\setcounter{section}{0}

\section{Part Segmentation}
\label{sec:shapenet_supp}

\section{Indoor Scene Segmentation}
\label{sec:s3dis_supp}

\end{document}